\documentclass[]{youtu} 
\PassOptionsToPackage{table}{xcolor}
\usepackage{CJKutf8}
\usepackage{mathpazo}
\usepackage{hyperref}
\usepackage{booktabs}
\usepackage{array}
\usepackage{url}
\usepackage[utf8]{inputenc} 
\usepackage[T1]{fontenc}      
\usepackage{wrapfig}
\usepackage{url}           
\usepackage{amsfonts}       
\usepackage{nicefrac}       
\usepackage{microtype}      
\usepackage[table]{xcolor}
\usepackage{bm}
\usepackage{makecell}
\usepackage{enumitem}
\usepackage{booktabs}
\usepackage{array}
\usepackage{multirow}
\usepackage{bbding}
\usepackage{colortbl}
\definecolor{mygray}{gray}{.92}
\usepackage[ruled,vlined]{algorithm2e}
 
\usepackage{amsmath,amsthm,amssymb}
\usepackage{tcolorbox}
\usepackage{color}
\usepackage{graphicx}
\usepackage{subcaption}
\usepackage{caption}
\usepackage{natbib}
\usepackage{algorithmic}
\usepackage{pifont} 
\newcommand{\cmark}{\ding{51}}
\newcommand{\xmark}{\ding{55}}

\title{SwiftVideo: A Unified Framework for Few-Step Video Generation through Trajectory-Distribution Alignment
}
\author{\small Yanxiao Sun\textsuperscript{$1 \dagger$}, Jiafu Wu\textsuperscript{$2 \dagger$}, Yun Cao\textsuperscript{$2$}, Chengming Xu\textsuperscript{$2$}, Yabiao Wang\textsuperscript{$2$},  \\Weijian Cao\textsuperscript{$2$}, Donghao Luo\textsuperscript{$2$}, Chengjie Wang\textsuperscript{$2$},Yanwei Fu\textsuperscript{$1,3$}}
\vspace{-5mm}
\affiliation{$^1$Fudan University\\\textsuperscript{$2$}Tencent Youtu Lab\\\textsuperscript{$3$}Shanghai Innovation Institute}

\youtufinalcopy

\begin{document}

\abstract{
Diffusion-based or flow-based models have achieved significant progress in video synthesis but require multiple iterative sampling steps, which incurs substantial computational overhead. While many distillation methods that are solely based on trajectory-preserving or distribution-matching have been developed to accelerate video generation models, these approaches often suffer from performance breakdown or increased artifacts under few-step settings. To address these limitations, we propose \textbf{\emph{SwiftVideo}}, a unified and stable distillation framework that combines the advantages of trajectory-preserving and distribution-matching strategies. 
Our approach introduces continuous-time consistency distillation to ensure precise preservation of ODE trajectories. Subsequently, we propose a dual-perspective alignment that includes distribution alignment between synthetic and real data along with trajectory alignment across different inference steps.
Our method maintains high-quality video generation while substantially reducing the number of inference steps. 
Quantitative evaluations on the OpenVid-1M benchmark demonstrate that our method significantly outperforms existing approaches in few-step video generation.  
}
\maketitle

\section{Introduction}
\renewcommand{\thefootnote}{}
\footnotetext{$\dagger$ Equal contribution.}
\vspace{-5mm}
Recent advancements in diffusion and flow-based models have significantly enhanced generative modeling, especially in video synthesis. However, these models typically require many iterative sampling steps, resulting in high computational and time costs that limit their practical deployment. Although higher-order numerical ODE solvers \citep{dpm++, dpm} can reduce the number of sampling steps, they often suffer from severe quality degradation when operating in the few-step regime (i.e., $\textnormal{steps}\leq10$). 
\vspace{-2mm}

To address these challenges, diffusion distillation methods have emerged to reduce sampling overhead in video generation. Trajectory-preserving approaches, such as LCM, PCM \citep{lcm, pcm}, leverage consistency loss to minimize inference steps.
However, as discrete-time consistency models (CMs), they tend to produce blurry outputs under few-step settings and introduce additional discretization errors from numerical ODE solvers.
Alternatively, distribution-matching methods like DMD2 \citep{dmd2}, train student models to approximate the teacher's distribution in order to reduce inference steps. However, this approach inherently upper-bounds the student model's performance by the teacher model and can lead to domain inconsistencies due to relying solely on distributional constraints. Other recent strategies \citep{hyper-sd, ladd} incorporate post-training enhancements for few-step generation, but often depend on auxiliary reward models or expensive-to-collect preference datasets.
\vspace{-2mm}

Based on this analysis, we identify three key limitations in existing methods: 1) Discretization errors from numerical ODE solvers in previous CMs; 2) Performance ceilings imposed by teacher model distributions; 3) Quality degradation under extremely few steps $\textnormal{steps}\leq4$.
To this end, we propose \textbf{\emph{SwiftVideo}}, a unified and stable distillation framework that systematically addresses these limitations. Our framework comprises three components: continuous-time consistency distillation (CCD) for precise ODE trajectory preservation, distribution alignment (DA) on real data to enhance visual fidelity, and trajectory alignment (TA) across different inference steps for enhanced few-step generation.

\begin{wrapfigure}{l}{0.5\linewidth} 
\vspace{-5mm}
    \centering
    \includegraphics[width=0.95\linewidth]{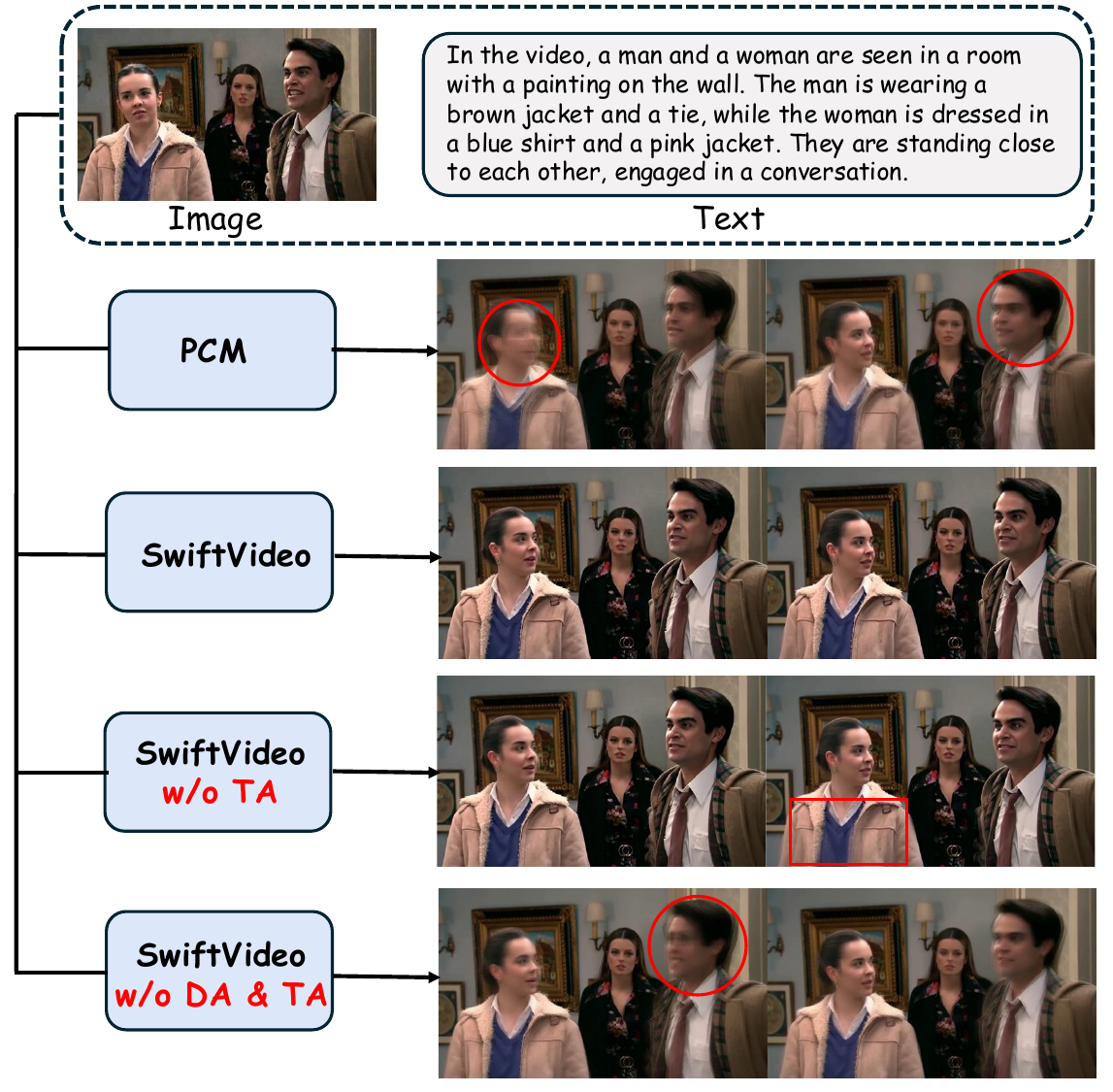}
    \caption{Video generation results on the OpenVid-1M dataset \citep{openvid} using 4 sampling steps. Red circles and boxes highlight blurry regions and artifacts in the generated videos. DA and TA denote Distribution Alignment and Trajectory Alignment, respectively. SwiftVideo w/o DA \& TA employs only Continuous-time Consistency Distillation (CCD). Our method significantly outperforms PCM.}
    \label{fig:teaser}
\end{wrapfigure}
Firstly,  we introduce CCD to ensure precise ODE trajectory preservation, eliminating the discretization errors that arise from ODE solver reliance in discrete-time consistency models. 
As shown in Figure \ref{fig:teaser}, while CCD alone outperforms PCM, it still exhibits blurriness. Nevertheless, its accurate trajectory preservation enables effective subsequent training optimization.
Secondly, instead of approximating the teacher's distribution, we directly perform DA on real data, thereby breaking the performance constraints imposed by the teacher model. 
Thirdly, we observe that the distribution-aligned distilled model exhibits significant improvement in generation quality as the number of steps increases within the low-step regime (i.e., $\text{steps}\leq8$). Based on this observation, we propose TA across different inference steps as a post-training strategy. 
Specifically, we generate the preference dataset using the distilled model with varying numbers of steps, then leverage the Direct Preference Optimization (DPO) algorithm \citep{dpo} to implicitly align low-step inference trajectories with high-step inference trajectories, eliminating the need for reward models and costly external preference data collection.
As shown in Figure \ref{fig:teaser}, DA effectively reduces blurriness while TA improves detail preservation under few-step settings.
\vspace{-2mm}

In summary, our main contributions are as follows:
(1) We propose a unified and stable distillation framework that maintains high-quality video generation while significantly reducing the number of inference steps. 
(2) \textbf{Continuous-time Consistency Distillation (CCD)}: We effectively apply CCD to video generation models for the first time, ensuring precise ODE trajectory preservation while incorporating stabilization and acceleration techniques throughout the training process.
(3) \textbf{Distribution Alignment (DA)}: We employ adversarial training to enable the model to directly align with real data distributions, thereby enhancing the visual details and realism of generated results.
(4) \textbf{Trajectory Alignment (TA)}: We apply the DPO algorithm on synthetic preference dataset across different inference steps to enforce alignment with high-step inference trajectories, thus enhancing few-step generation quality.
(5) Extensive experiments demonstrate that our approach outperforms existing methods in few-step video generation, achieving state-of-the-art results on VBench and FVD benchmarks.
\vspace{-3mm}

\section{Related Work}
\vspace{-2mm}
\subsection{Video Diffusion Model}
\vspace{-2mm}
Driven by advances in large-scale video data and model architectures, video diffusion models have achieved remarkable success. Early approaches \citep{svd, animatediff}  using 3D U-Net architectures in latent space were limited to generating short, low-quality videos. With the success of Sora \citep{sora}, modern models have shifted toward Diffusion Transformer (DiT) architectures, with open-source implementations like HunyuanVideo \citep{hunyuanvideo}, CongVideoX \citep{cogvideox}, and Wan \citep{wan}. Although these models can produce longer, higher-quality videos, generation remains computationally expensive and time-consuming. To address this limitation, we leverage diffusion distillation methods to accelerate these models without compromising their quality.
\vspace{-2mm}

\subsection{Diffusion Model Distillation}
Diffusion models typically rely on iterative inference to generate high-quality samples, which is both time-consuming and computationally expensive. Diffusion model distillation has been regarded as an effective technique for accelerating the inference process and can be categorized into trajectory-preserving and distribution-matching approaches.
\vspace{-2mm}

Trajectory-preserving distillation aims to train a student model capable of generating samples in fewer steps by approximating the ordinary differential equation (ODE) sampling trajectory of the teacher model. 
Progressive Distillation \citep{progressive} trains the student model to predict the subsequent flow locations determined by the teacher model. 
Rectified Flow \citep{rectified_flow} trains the student model on noisy-image pairs obtained from the teacher, thereby progressively straightening the ODE trajectory. 
Consistency Distillation \citep{lcm} trains a student model to map noisy samples along the ODE trajectory to its origin. Although trajectory-preserving distillation facilitates training, the samples generated in few-step tend to exhibit blurriness and inferior quality.
\vspace{-2mm}

Distribution-matching distillation trains a student model to approximate the teacher's distribution. Some methods leverage adversarial loss functions to improve the generative quality of student models \citep{add, ladd}. DMD \citep{dmd} minimizes approximate KL divergence using gradients from two score functions: a fixed pretrained target score and a dynamically trained generator score.
\vspace{-2mm}

Recently, trajectory-preserving distillation methods \citep{t2v-turbo, videolcm} and  distribution-matching distillation\citep{sf-v}) have been increasingly applied to video generative models. However, many of these distilled video models originate from the pre-Sora era and typically generate only low-quality, short-duration results. Moreover, existing consistency distillation methods for video are trained using discretized timesteps, which rely on numerical ODE solvers and introduce additional discretization errors. In contrast, our approach employs continuous-time consistency distillation to address these issues. Furthermore, prior distribution-matching distillation methods usually approximate the teacher’s output distribution as the target, which implicitly limits the generation quality by the teacher model. Instead, our approach directly performs distribution alignment on real data, similar to \citep{apt}.
\vspace{-2mm}

\begin{figure*}[t]
    \centering
    \includegraphics[width=\linewidth]{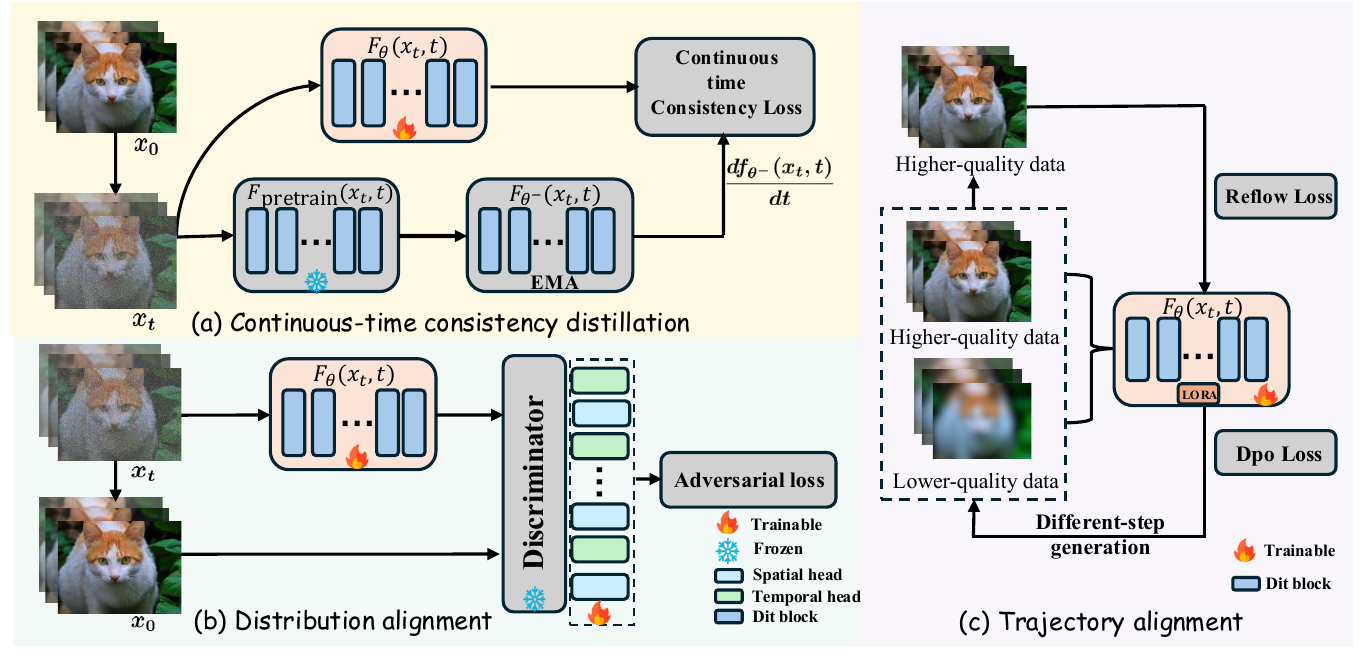}
    \caption{
The overview of our distillation framework SwiftVideo.
}
    \label{fig:framework}
\end{figure*}

\section{Preliminary}
\vspace{-2mm}
\subsection{Flow Matching}
\vspace{-2mm}
Flow Matching \citep{flowmatching, rectified_flow} is widely used in recent advanced video-generation models \citep{wan}. Given a  data sample $\mathbf{x}_0\sim p_{\textnormal{data}}$ and a noise sample $\mathbf{x}_1 \sim p_{\textnormal{noise}}$, the noisy data $\mathbf{x}_t$ is defined as a linear interpolation between these two distributions:
\begin{equation}
    \mathbf{x}_t=(1-t)\mathbf{x}_0+t\mathbf{x}_1,
\end{equation}
for $t\in[0,1]$. 
The target velocity field is $\boldsymbol{v}=\mathbf{x}_1-\mathbf{{x}_0}$. The optimization process of flow matching is defined as:
\begin{equation}
    \mathcal{L}(\theta)=\mathbb{E}_{t,\mathbf{x}_0\mathbf{\sim}X_0,\mathbf{x}_1\mathbf{\sim}X_1}\left[\|\boldsymbol{v}-\boldsymbol{v}_\theta(\mathbf{x}_t,t)\|^2\right],
\end{equation}
where $\boldsymbol{v}_{\theta}$ denotes the model parameterized by $\theta$.
\vspace{-2mm}
\subsection{Consistency Models}
\vspace{-2mm}
A consistency model (CM) is a neural network $\bm{f}_\theta(\mathbf{x}_t,t)$ trained to map any points on a trajectory of the PF-ODE to the trajectory’s origin. The formula can be described as $\bm{f}:(\mathbf{x}_t,t)\mapsto \mathbf{x}_0$. A valid $f_\theta$ must satisfy the boundary condition $\mathbf{f}_\theta(\mathbf{x}, 0)\equiv\mathbf{x}$, which can be achieved through the parameterization $\bm{f}_\theta(\mathbf{x}_t,t)=c_\text{skip}(t)\mathbf{x}_t+c_\text{out}(t)\bm{F}_\theta(c_\text{in}(t)\mathbf{x}_t, c_\text{noise}(t))$, where $c_\text{skip}(0)=1$ and $c_\text{out}(0)=0$. Depending on the selection of nearby time steps, consistency models can be classified into two categories, as described below:
\vspace{-2mm}
\subsubsection{Discrete-time CMs}
\vspace{-2mm}
The training objective defined on two adjacent time steps with a finite distance is as follows:
\begin{equation}
\mathbb{E}_{\mathbf{x}_t,t}\left[w(t)d(\bm{f}_\theta(\mathbf{x}_t,t),\bm{f}_{\theta^-}(\mathbf{x}_{t\mathbf{-}\Delta t},t\mathbf{-}\Delta t))\right]
\label{eq:cm}
\end{equation}
where $\theta^{-}$ denotes $stopgrad(\theta)$, which is updated with exponential moving average (EMA) of the parameter $\theta$. The term $w(t)$ is a weighting function, $\Delta t > 0$ is the distance between adjacent timesteps, and $d(\cdot, \cdot)$ is a distance function, e.g., the squared $\ell_2$ distance $d(\mathbf{x},\mathbf{y})=||\mathbf{x}-\mathbf{y}||^{2}_{2}$. The noisy sample $x_{t-\Delta t}$ at the preceding time step $t-\Delta t$ is typically obtained by solving the PF-ODE using numerical ODE solvers with a step size of $\Delta t$, which introduce additional discretization error.
\vspace{-2mm}
\subsubsection{Continuous-time CMs}
\vspace{-2mm}
When taking the limit $\Delta t\to0$, show that the gradient of Eq (\ref{eq:cm}) with respect to $\theta$ converges to
\begin{equation}
    \label{eq:scm}
    \nabla_\theta {E}_{\mathbf{x}_t, t}\left[ w(t) \bm{f}_\theta^\top(\mathbf{x}_t, t) \frac{\mathrm{d} \bm{f}_{\theta^-}(\mathbf{x}_t, t)}{\mathrm{d}t} \right],
\end{equation}
where $\frac{\mathrm{d} \bm{f}_{\theta^-}(\mathbf{x}_t, t)}{\mathrm{d}t} = \nabla_{x_t} \bm{f}_{\theta^-}(\mathbf{x}_t, t) \frac{\mathrm{d} \mathbf{x}_t}{\mathrm{d}t} + \partial_t \bm{f}_{\theta^-}(\mathbf{x}_t, t)$ is the tangent of $\bm{f}_{\theta^-}$ at $(\mathbf{x}_t, t)$ along the trajectory of the PF-ODE $\frac{\mathrm{d}\mathbf{x}_t}{\mathrm{d}t}$. Continuous-time CMs do not rely on ODE solvers, which  avoids discretization errors and offers more accurate supervision signals during training. Recently, sCM \citep{scm} presents several training tricks to simplify, stabilize, and scale the training of continuous-time consistency models, achieving superior performance compared to discrete-time consistency models.
\vspace{-2mm}

\section{Method}
\vspace{-2mm}
As shown in Figure \ref{fig:framework}, our framework systematically addresses the three limitations identified above through the following components: continuous-time consistency distillation, distribution alignment via adversarial training, and DPO-based trajectory alignment with reflow loss regularization.
\vspace{-2mm}
\subsection{Continuous-time Consistency Distillation}
\vspace{-2mm}
\begin{algorithm}[tb]
\caption{Continuous-time Consistency Distillation}
\label{alg:distill}
\begin{algorithmic}[1] 
\STATE \textbf{Input}: Training dataset $D$,  pretrained model $\bm{F}_{\textnormal{pretrain}}$, distilled model $\bm{F}_{\theta}$, learning rate $\eta$, ema decay $\mu$, timestep schedule $(P_{\textnormal{mean}},P_{std}^{2})$, consistency loss warmup iteration $H$ \\
\STATE \textbf{Initialize} $\textnormal{Iters} \gets 0$, $\theta \gets \phi$
\REPEAT
    \STATE $\mathbf{x}_0 \sim \mathcal{D}$, $\mathbf{x}_1\sim\mathcal{N}(0,\mathbf{I})$, $t\sim\mathcal{N}(P_{\textnormal{mean}},P_{\textnormal{std}}^2)$
    \STATE $\mathbf{x}_t\gets(1-t)\mathbf{x}_0+t\mathbf{x}_1$, $\frac{\mathrm{d}\mathbf{x}_t}{\mathrm{d}t}\gets\bm{F}_\text{pretrain}(\mathbf{x}_t,t)$
    \STATE $r \gets \textnormal{min(1, \textnormal{Iters}/H})$
    \STATE $\mathbf{g \gets}\frac{\mathrm{d}\mathbf{x}_t}{\mathrm{d}t}-\bm{F}_{\theta^-}(\mathbf{x}_t,t)-rt\cdot\mathbf{sg}(\frac{\mathrm{d}\bm{F}_{\theta^-}(\mathbf{x}_t,t)}{\mathrm{d}t})$
    \STATE $\mathbf{g} \leftarrow \mathbf{g} / (\|\mathbf{g}\| + c)$
    \STATE $\mathcal{L}_{\textnormal{distill}} \leftarrow \left\| \bm{F}_\theta\left( \mathbf{x}_t, t \right) - \mathbf{F}_{\theta^{-}}\left( \mathbf{x}_t, t\right) - \mathbf{g} \right\|_2^2$
    \STATE $\theta \gets \theta-\eta \nabla_{\theta} \mathcal{L}_\textnormal{distill}$
    \STATE $\theta^{-}\gets\mathbf{sg}\left(\mu\theta^{-}+(1-\mu)\theta\right)$
    \STATE $\textnormal{Iters} \gets\textnormal{Iters}+1$
    
\UNTIL{convergence}

\end{algorithmic}
\end{algorithm}
In previous work, the consistency model $\bm{f}_{\theta}$ is parameterized as:
\begin{equation}
    \bm{f}_\theta(\mathbf{x}_t,t)=c_\mathrm{skip}(t)\mathbf{x}_t+c_\mathrm{out}(t)\bm{F}_\theta(\mathbf{x}_t,t)
\end{equation}
$\bm{F}_{\theta}(x,t)$ is a deep neural network.
Unlike the TrigFlow formulation employed in sCM \citep{scm}, we set $c_\mathrm{skip}=1$ and $c_\mathrm{out}=-t$, which aligns with the Rectified Flow \citep{rectified_flow} formulation commonly adopted in current video diffusion models and eliminates the need for additional formulation conversion. Moreover, this parameterization satisfies the boundary conditions and is compatible with the Euler ODE solver. Then we differentiate both sides of Eq (\ref{eq:scm}) with respect to $t$:
\begin{equation}
\frac{\mathrm{d} \bm{f}_{\theta^-}(\mathbf{x}_t, t)}{\mathrm{d}t}=\frac{\mathrm{d}\mathbf{x}_t}{\mathrm{d}t}-\bm{F}_{\theta^-}(\mathbf{x}_t,t)-t\frac{\mathrm{d}\bm{F}_{\theta^-}(\mathbf{x}_t,t)}{\mathrm{d}t}
\label{eq:dfdt}
\end{equation}
where $\theta^{-}$ is updated with exponential moving average (EMA) of the paramenter $\theta$. In this work, we aim to distill a pretrained video generation model into a few-step generator, so $\frac{\mathrm{d}\mathbf{x}_t}{\mathrm{d}t}=\bm{F}_\text{pretrain}(\mathbf{x}_t,t)$ in Eq. (\ref{eq:dfdt}).
\vspace{-2mm}

Given the equation $\nabla_\theta\mathrm{E}[\bm{F}_\theta^\top\mathbf{y}]=\frac{1}{2}\nabla_\theta\mathrm{E}[\|\bm{F}_\theta-\bm{F}_{\theta^-}+\mathbf{y}\|_{2}^{2}$, where $\mathbf{y}$ is an arbitrary vector independent of $\theta$. Then we can derive the training objective for continuous-time consistency distillation:
\begin{equation}
    \begin{aligned}
      &\nabla_\theta \mathrm{E}_{\mathbf{x}_t, t}\left[\bm{f}_\theta^\top(\mathbf{x}_t, t) \frac{\mathrm{d} \bm{f}_{\theta^-}(\mathbf{x}_t, t)}{\mathrm{d}t} \right]\\
      &=-t\nabla_\theta \mathrm{E}_{\mathbf{x}_t, t}\left[\bm{F}_\theta^\top(x_t, t) \frac{\mathrm{d} \bm{f}_{\theta^-}(\mathbf{x}_t, t)}{\mathrm{d}t} \right]\\
      &=\frac{t}{2}\nabla_\theta \mathrm{E}_{\mathbf{x}_t, t}\|\bm{F}_\theta(\mathbf{x}_t, t)-
      \bm{F}_{\theta^-}(\mathbf{x}_t, t)-\frac{\mathrm{d} \bm{f}_{\theta^-}(\mathbf{x}_t, t)}{\mathrm{d}t}\|_2^2\\
    \end{aligned}
    \label{eq:continuous-time}
\end{equation}
However, directly training on this objective suffers from instability and inefficiency issues,  it's necessary to incorporate additional training techniques for stabilization and acceleration.
\vspace{-2mm}

\subsubsection{Training Stabilization and Acceleration} 
\vspace{-2mm}
Building upon the approach proposed in sCM \citep{scm}, we employed two training techniques, Tangent Normalization and Tangent Warmup, to mitigate the instability issues in the continuous-time consistency distillation process. Additionally, given that current video diffusion models comprise multiple DiT blocks, we designed an iterative algorithm to compute the tangent function $\frac{\mathrm{d} \bm{f}_{\theta^-}(\mathbf{x}_t, t)}{\mathrm{d}t}$ with minimal computational overhead.
\vspace{-2mm}

In our experiments, we found that the term $\frac{\mathrm{d}\bm{F}_{\theta^-}(\mathbf{x}_t,t)}{\mathrm{d}t}$ within the tangent function $\frac{\mathrm{d} \bm{f}_{\theta^-}(\mathbf{x}_t, t)}{\mathrm{d}t}$ is the primary cause of instability in the training objective of continuous-time Flow Distillation. This term can be derived as follows using the chain rule:
\begin{equation}
    \frac{\mathrm{d}\bm{F}_{\theta^-}(\mathbf{x}_t,t)}{\mathrm{d}t}=\nabla_{\mathbf{x}_t}\bm{F}_{\theta^-}(\mathbf{x}_t,t)\frac{\mathrm{d}\mathbf{x}_t}{\mathrm{d}t}+\partial_t\bm{F}_{\theta^-}(\mathbf{x}_t,t)
\end{equation}
\vspace{-2mm}

Firstly, we employed the Tangent Warmup training trick by introducing a coefficient $r$ to scale the term $t\frac{\mathrm{d}\bm{F}_{\theta^-}(\mathbf{x}_t,t)}{\mathrm{d}t}$ in the tangent function Eq.(\ref{eq:dfdt}). This coefficient $r$ linearly increases from 0 to 1 over the first 1K training steps. Thus, the tangent function is transformed into the following form.
\begin{equation}
    \begin{aligned}
    &\frac{\mathrm{d}\mathbf{x}_t}{\mathrm{d}t}-\bm{F}_{\theta^-}-rt\frac{\mathrm{d}\bm{F}_{\theta^-}(\mathbf{x}_t,t)}{\mathrm{d}t}\\
    &=(1-r)(\frac{\mathrm{d}\mathbf{x}_t}{\mathrm{d}t}-\bm{F}_{\theta^-})+r(\frac{\mathrm{d}\mathbf{x}_t}{\mathrm{d}t}-\bm{F}_{\theta^-}-t\frac{\mathrm{d}\bm{F}_{\theta^-}(\mathbf{x}_t,t)}{\mathrm{d}t})
    \label{eq:dfdt_r}
    \end{aligned}
\end{equation}
We observe that Eq.(\ref{eq:dfdt_r}) is expressed as a weighted sum of flow matching loss and original training objective of continuous-time consistency distillation in Eq.(\ref{eq:continuous-time}). 
\vspace{-2mm}

Secondly, We employed the Tangent Normalization to reduce the variance during training, experimenting with two approaches: clipping the tangent within the range of $-1$ to $1$ versus  replacing $\frac{\mathrm{d}\bm{f}_{\theta^-}(\mathbf{x}_t,t)}{\mathrm{d}t}$ with $\frac{\mathrm{d}\bm{f}_{\theta^-}(\mathbf{x}_t,t)}{\mathrm{d}t}/(\left\|\frac{\mathrm{d}\bm{f}_{\theta^{-}}(\mathbf{x}_t,t)}{\mathrm{d}t}\right\|+c),c=0.1$. We found that the latter approach yielded better results.
\vspace{-2mm}

While the above two training tricks stabilize the training process, the computation of $\frac{\mathrm{d}\bm{F}_{\theta^-}(\mathbf{x}_t,t)}{\mathrm{d}t}$ that lead to efficiency and memory issues when directly applied to video diffusion models. To address this, we apply stop-gradient operations in JVP computation and perform iterative block-wise JVP calculation across DiT blocks, using the JVP output from each block as input for the subsequent block's JVP computation while detaching intermediate results to manage memory consumption. The detailed process for JVP computation is described in the appendix.
\vspace{-2mm}

Algorithm \ref{alg:distill} presents our continuous-time consistency distillation algorithm.
\vspace{-2mm}

\subsection{Alignment for Distilled Model}
\vspace{-2mm}
To further improve generation quality in few-step sampling, we align the distilled model from both distribution and trajectory perspectives.
\vspace{-2mm}
\subsubsection{Distribution Alignment}
\vspace{-2mm}

While continuous-time consistency distillation achieves precise ODE trajectory preservation, the generated results still exhibit blurriness due to the inherent properties of consistency models \citep{sdxl-lightning}. Therefore, we introduce distribution alignment to improve the visual fidelity and realism of generated videos.
Unlike previous distribution-matching methods that constrain the upper limit of the student's generation quality by matching the teacher's distribution, we directly employ adversarial loss to align with real data distribution.
\vspace{-2mm}

The distribution alignment employs a discriminator trained to distinguish between generated samples $\hat{\mathbf{x}}_0$ and real data $\mathbf{x}_{0}$, where $\hat{\mathbf{x}}_0=\mathbf{x}_t-t\bm{F}_{\theta}(\mathbf{x}_t,t)$.
For the discriminator, we follow the design in StyleGAN-T \citep{stylegan-t}. We utilize a frozen pretrained feature network DINOv2 \citep{dinov2} coupled with trainable lightweight discriminator heads, comprising both temporal heads for capturing inter-frame temporal dynamics and spatial heads for evaluating intra-frame visual quality.
\vspace{-2mm}

For training stability, we introduce distribution alignment after $\mathcal{N}_{\textnormal{warmup}}$ training steps of continuous-time consistency distillation. The objective of the distribution alignment training is as follows:
\vspace{-2mm}
\begin{equation}
\begin{aligned}
    \mathcal{L}^\mathcal{G}&=\lambda_{\textnormal{adv}} \cdot\mathbb{E}_{\mathbf{x}_0,t}\left[\textstyle\sum_k\textnormal{max}(0, 1-\mathcal{D}_{\phi,k}(\hat{\mathbf{x}}_0))\right]+\mathcal{L}_{\mathrm{distill}}\\
    \mathcal{L}^\mathcal{D}&=\mathbb{E}_{\mathbf{x}_0,t}\left[\textstyle\sum_k\textnormal{max}(0,1-\mathcal{D}_{\phi,k}(\mathbf{x}_0)), \textnormal{max}(0, 1+\mathcal{D}_{\phi,k}(\hat{\mathbf{x}}_0))\right]
\end{aligned}
\end{equation}
where $\lambda_{\textnormal{adv}}$ is a hyper-parameter, and $\mathcal{D}_{\phi,k}$ denotes either spatial or temporal discriminator heads.
\vspace{-2mm}

By directly aligning distributions through adversarial training on real data, our approach significantly enhances visual fidelity and realism in video generation.
\vspace{-2mm}

\subsubsection{Trajectory Alignment}
\begin{wrapfigure}{l}{0.5\linewidth} 
\vspace{-5mm}
    \centering
    \includegraphics[width=0.95\linewidth]{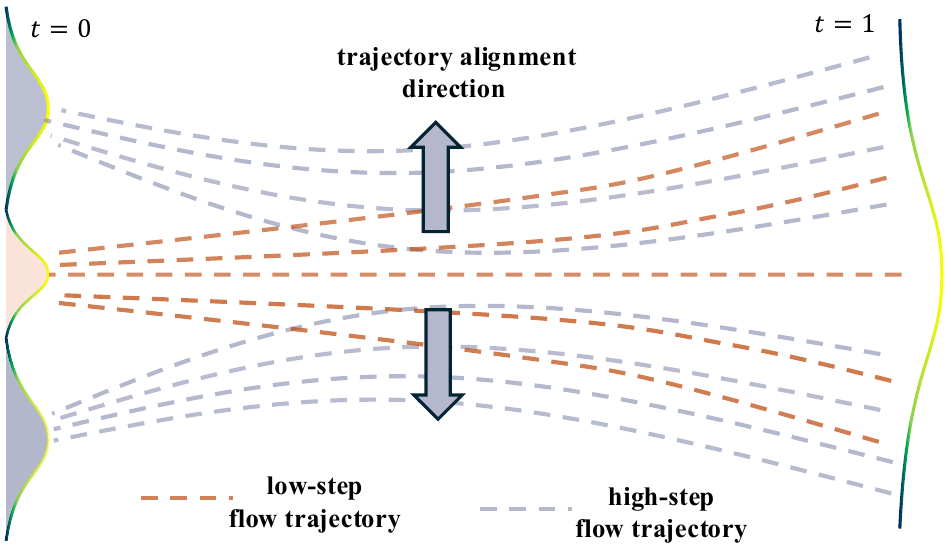}
    \caption{Illustration of trajectory alignmant.}
    \label{fig:ta}
\end{wrapfigure}
While distribution alignment enhances the generation quality of distilled models under limited inference steps, visual artifacts persist at extremely low step counts (i.e., $\textnormal{steps} \leq 4$). We observe that distilled model using Euler solver inference exhibits significant quality improvements as the number of steps increases within the low-step regime (i.e., $\textnormal{steps} \leq 8$). Based on this observation, we introduce trajectory alignment across different inference steps to advance model's low-step generation frontier.
Initially, we utilize the distribution-aligned distilled model to efficiently construct a synthetic preference dataset $\mathcal{D}=\{\mathbf{y},\mathbf{x}_0^w,\mathbf{x}_0^l\}$ across varying inference steps. In this preference dataset, each sample contains a prompt $\mathbf{y}$ and two videos $\mathbf{x}_0^w$
and $\mathbf{x}_0^l$ generated by the  distribution-aligned distilled model using different steps (e.g., 8 steps and 4 steps) respectively.
\vspace{-2mm}

Subsequently, we finetune our model using Direct Preference Optimization \citep{flow-dpo} on this synthetic preference dataset. 
As shown in Figure \ref{fig:ta}, since the synthetic data implicitly contains flow trajectory information, this training enables the model to align with high-step inference trajectories, thereby enhancing low-step generation quality.
The DPO loss $\mathcal{L}_{\textnormal{DPO}}$ on flow-based video generation model is written below:
\begin{equation}
\begin{aligned}
    \mathbf{-}\mathbb{E}\bigg[\log\sigma\bigg(-&\frac{\beta}2\bigg(\|\mathbf{v}^w-\mathbf{v}_\theta(\mathbf{x}_t^w,t)\|^2-\|\mathbf{v}^w-\mathbf{v}_\mathrm{ref}(\mathbf{x}_t^w,t)\|^2\\&-\left(\|\mathbf{v}^l-\mathbf{v}_\theta(\mathbf{x}_t^l,t)\|^2-\|\mathbf{v}^l-\mathbf{v}_\mathrm{ref}(\mathbf{x}_t^l,t)\|^2)\right)\bigg],
\end{aligned}
\label{flow-dpo}
\end{equation}
where $\beta$ is a hyperparameter, $\{\mathbf{x}_0^w,\mathbf{x}_0^l\}\sim\mathcal{D}$ and $\mathbf{x}_t^*=(1-t)\mathbf{x}_0^*+t\mathbf{\epsilon},\mathbf{\epsilon}\sim\mathcal{N}(0,\mathbf{I})$, with $*$ denoting either $w$ (for the preferred data) or $l$ (for the less preferred data). 
\vspace{-2mm}

Employing Eq. (\ref{flow-dpo}) alone as the training objective for trajectory alignment results in undesired training dynamics. As training progresses, the model minimizes the loss by reducing velocity prediction accuracy on dispreferred data more than on preferred data, thereby causing the flow to diverge from the intended trajectory. To address this issue, we introduce a regularization term $\|\mathbf{v}^w-\mathbf{v}_\theta(\mathbf{x}_t^w,t)\|^2$ to maintain high prediction fidelity on the preferred data. Given that our preference dataset is entirely synthetic, this regularization term takes the same form as the reflow loss \citep{rectified_flow}, denoted as $\mathcal{L}_{RF}$. Therefore, the overall trajectory alignment loss can be formulated as $\mathcal{L}_{\textnormal{DPO}}+\lambda_{\textnormal{RF}}\cdot\mathcal{L}_{\textnormal{RF}}$, where $\lambda_{\textnormal{RF}}$ is a hyper-parameter. 
We note that the model performance is robust to the choice of $\lambda_{\textnormal{RF}}$, as detailed in the appendix.
\vspace{-2mm}

\section{Experiment}
\vspace{-2mm}

\subsection{Implementation Details.} 
\vspace{-2mm}
In our distillation experiments, we use Wan2.1-FUN-inp-480p-1.3B \citep{wan} as the foundational model. For fair comparison, all the experiments are conducted on OpenVid-1M \citep{openvid}, a large-scale, high-quality video dataset that provides a standardized benchmark for few-step video generation. Specifically, we randomly select one million videos for training and 1000 for evaluation. In the continuous-time consistency distillation and distribution alignment, we fix the resolution of the training videos to $832\times480$ with 61 frames and a batch size of 1. We conduct full-parameter training for 3,000 iterations using a learning rate of $1\mathrm{e}-6$ for both the discriminator and student model. The default ema rate is 0.95, warm-up iteration $\mathcal{N}_{\textnormal{warmup}}$ is 1000  and $\lambda_{\textnormal{adv}}$ is set to 0.01.
In trajectory alignment, we first utilize distribution-aligned distilled model to generate videos using 4 steps and 8 steps sampling respectively, then construct preference dataset of size 5000. Then we apply LoRA \citep{lora} to fine-tune the model's linear layers on this preference dataset, using a learning rate of $1\mathrm{e}-6$ for approximately 2000 iterations. To further improve 2-step generation quality, we conduct an additional round of trajectory alignment, employing the once-aligned model to generate preference data using 2-step and 4-step sampling with the same training procedure. $\lambda_{RF}$ is set to 2 and $\beta$ in $\mathcal{L}_{\mathrm{DPO}}$ is set to 2500. All training uses the AdamW optimizer \citep{adamw}, and we employ the Euler solver for sampling.
\vspace{-2mm}
\subsection{Evaluation Metrics}
\vspace{-2mm}
Following existing work \citep{sf-v, causvid}, we use FVD \citep{fvd} and standard VBench-I2V benchmark \citep{vbench} to evaluate our model. FVD measures the $\text{Fr\'{e}chet distance}$ between feature distributions of real and generated videos. VBench is a comprehensive benchmark that is widely used to evaluate the video quality. 
All evaluations are conducted on the OpenVid-1M test set.
\vspace{-2mm}

\subsection{Comparisons on Image-to-Video Generation}
\vspace{-2mm}
\subsubsection{Quantitative Comparisons.}
\vspace{-2mm}
\begin{table}[t]
\centering

  \begin{tabular}{c|c|c|cc|cc}
    \toprule
     \multirow{2}{*}{Method} & \multirow{2}{*}{Step} &  \multirow{2}{*}{FVD$\downarrow$} & \multicolumn{4}{c}{{VBench}} \\ \cline{4-7} & & & \makecell{Temporal\\Quality} & \makecell{Frame\\Quality} &  \makecell{I2V\\Subject} & \makecell{I2V\\Background} \\
     \midrule
    \multirow{3}{*}{Teacher} 
    & 25 & 98.06 & 86.83 & 61.92  & 98.72 & 98.12\\
    & 8 & 150.82 & 84.95 & 58.76 & 98.68 & 98.10\\
    & 4 & 321.95 & 81.72 & 51.96  & 98.59 & 98.04\\
    \midrule
    LCM & 4 & 192.93 & 83.35 & 57.51 & 98.61 & 97.07 \\
    LCM & 2 & 591.51 & 79.6 & 45.9 & 98.45 & 97.94 \\
    \midrule
    PCM	& 4 & 198.56 & 83.23 &	57.29 & 98.61 & 97.05 \\
    \midrule
    DMD2 & 4 &	167.42 &  83.38&	57.35 &	98.63 & 98.08 \\
    \midrule
    $\textnormal{OSV}^\dagger$ & 4 & 146.59 &	83.97 &	59.79 & 98.75 & 98.14\\
    $\textnormal{OSV}^\dagger$ & 2 & 471.38 & 80.41 & 49.09 & 98.55 & 98.01\\
   \midrule
    \textbf{Ours} & 4 &	\textbf{118.89} & \textbf{85.27} & \textbf{61.76}  & \textbf{98.82} & \textbf{98.20} \\
    \textbf{Ours} & 2 & \textbf{199.61} & \textbf{83.17} & \textbf{59.16} & \textbf{98.77} & \textbf{98.14}\\
    \bottomrule
    \end{tabular}
\caption{Image-to-Video quantitative results on OpenVid-1M test set. FVD is computed using 64 frames. $\dagger$ donates our implementations.}
\label{tab:i2v}
\end{table}
We conduct comprehensive evaluations by comparing our model against state-of-the-art distillation methods, including trajectory-preserving methods LCM \citep{lcm} and PCM \citep{pcm}, and distribution-matching methods OSV \citep{osv} and DMD2 \citep{dmd2}.
We implement these methods on Wan 2.1 following the official codebases of LCM, PCM, and DMD2. And we try our best to implement OSV, which we denote as $\textnormal{OSV}^\dagger$. 
During evaluation, we utilize the first frame from the Openvid-1M test set as conditional input to generate 1000 videos at 832×480 resolution with 24 FPS, then compute FVD and VBench metrics on these generated videos. 

As shown in Table \ref{tab:i2v}, our method significantly outperforms existing  state-of-the-art distillation methods across all these quantitative metrics. Specifically, our approach achieves state-of-the-art FVD scores of 118.89 with 4-step inference and 199.61 with 2-step inference, indicating that our generated videos have the closest distribution to the ground truth. In VBench evaluations, our method with 2-step inference outperforms the 4-step performance of LCM, PCM, and DMD2 in terms of frame quality. Additionally, our method with 4-step inference approaches the teacher model's 25-step performance in frame quality, while even surpassing it in image-conditioning scores, indicating the capability of our method to preserve high-quality video synthesis under few-step inference. 

\subsubsection{Qualitative Comparisons}
\begin{figure*}[t]
    \centering
    \includegraphics[width=0.95\linewidth]{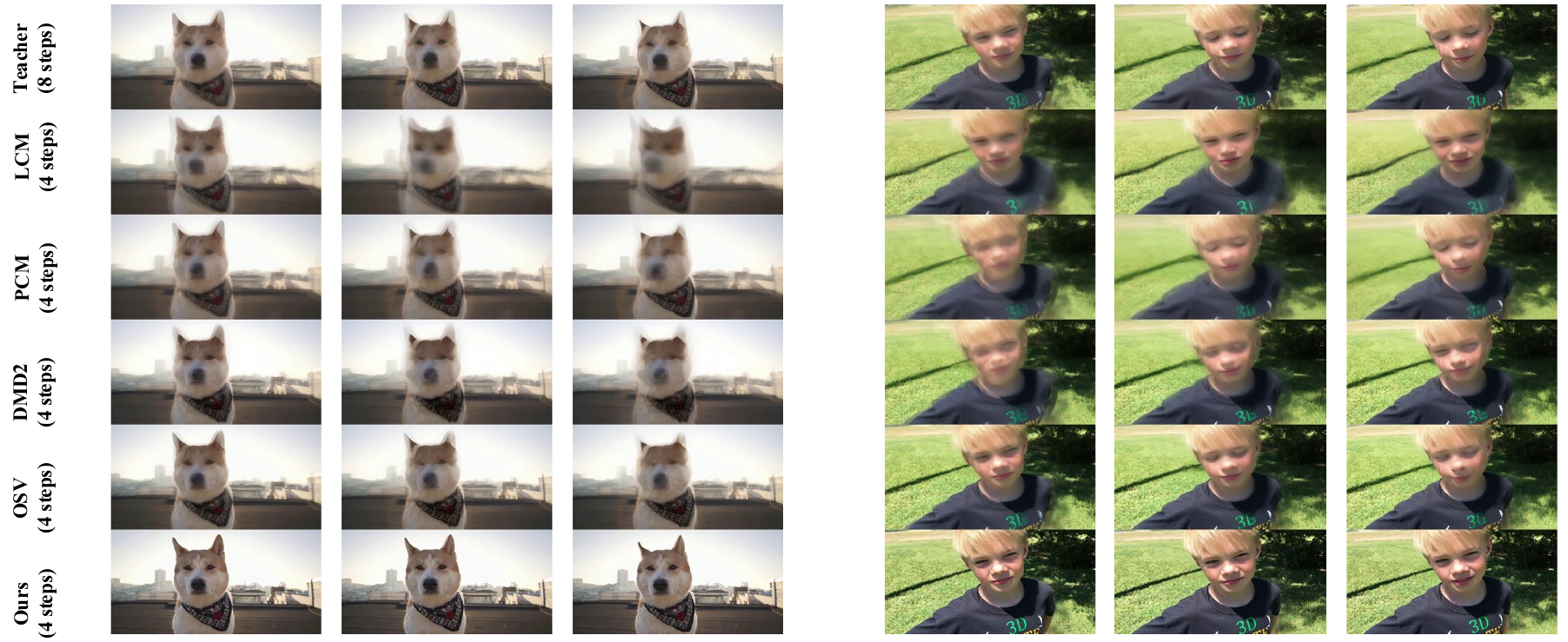}
    \caption{Qualitative comparison results. Our results significantly outperform other methods in both frame quality and temporal consistency under 4 sampling steps. All visual results are generated using images from the OpenVid-1M \citep{openvid} test set.}
    \label{fig:method_compare}
\end{figure*}

\vspace{-2mm}
Figure \ref{fig:method_compare} presents visual comparison results of the teacher model with 8 sampling steps, LCM \citep{lcm}, PCM \citep{pcm}, OSV \citep{osv}, DMD2 \citep{dmd2} and our method with 4 inference steps. Figure \ref{fig:method_compare} demonstrates that our method generates videos with superior visual clarity, richer details, and enhanced realism compared to other approaches, while exhibiting fewer artifacts. Additional qualitative results are provided in the appendix.
\vspace{-2mm}

\begin{figure*}[t]
    \centering
    \begin{subfigure}[t]{0.48\textwidth}
        \centering
        \includegraphics[width=\linewidth]{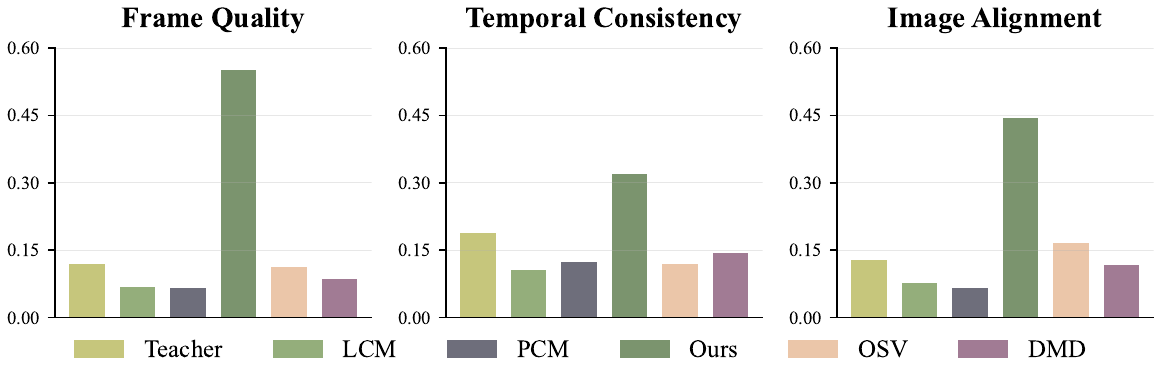}
        \caption{Results from our user study regarding frame quality, temporal consistency, and image alignment of different models. The teacher method uses 8 sampling steps, while other methods use 4 steps.}
        \label{fig:user_study}
    \end{subfigure}
    \hfill
    \begin{subfigure}[t]{0.48\textwidth}
        \centering
        \includegraphics[width=\linewidth]{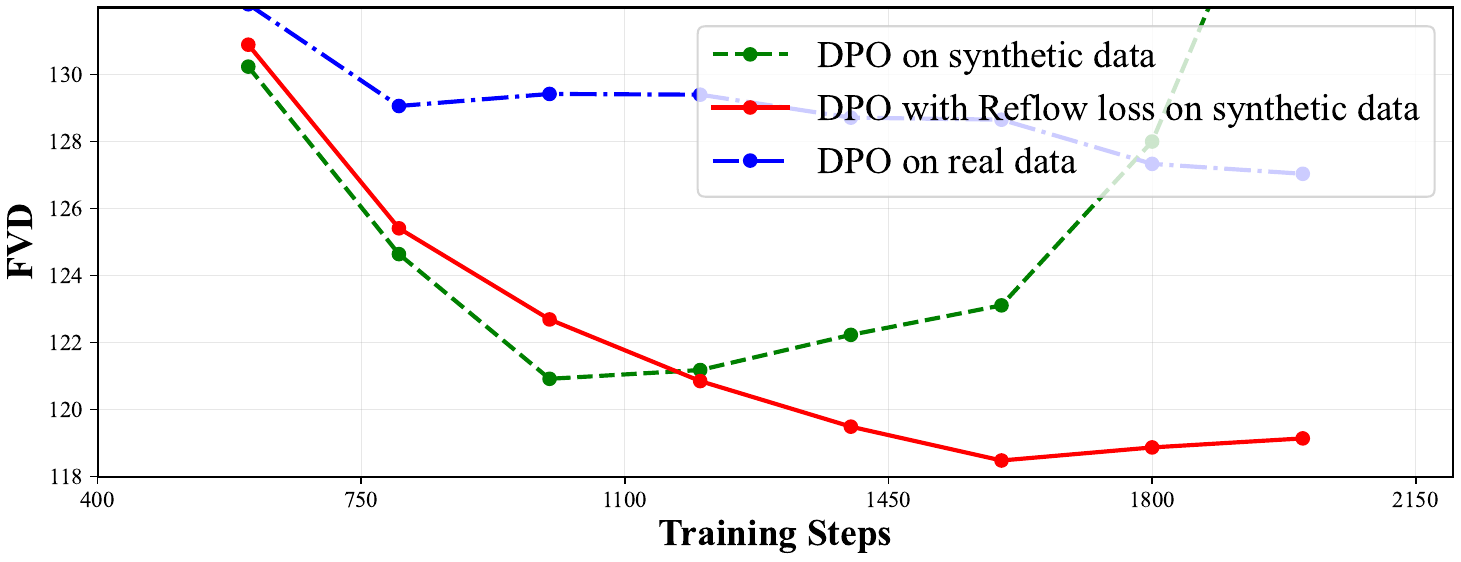}
        \caption{Effect of synthetic preference dataset and reflow loss in trajectory alignment.}
        \label{fig:ablation_dpo}
    \end{subfigure}
\end{figure*}

To further evaluate the performance of our method, we conduct a comprehensive user study comparing it against existing distillation approaches and the teacher model. We select first frames from 50 videos across different categories in the OpenVid-1M test set as conditional inputs. For each image, we generate 5 videos using same random seeds, with distillation methods using 4 sampling steps and the teacher model using 8 steps. Participants evaluate and select the best video based on frame quality, temporal consistency, and image alignment. As shown in Figure \ref{fig:user_study}, our approach significantly outperforms baseline distillation methods and the teacher model with 8 sampling steps.
\vspace{-2mm}

\begin{table}[t]
\label{tab:all}
\centering
\small
\begin{minipage}[c]{0.32\textwidth}
\subfloat[]{
\centering
\setlength{\tabcolsep}{3.0mm}
\begin{tabular}[t]{cc}
\toprule
Train setting & FVD$\downarrow$ \\	\midrule
$\textnormal{DA}$   &    163.54 \\  
$\textnormal{DCD}+\textnormal{DA}$   & 151.23 \\
$\textnormal{CCD}$   & 172.54  \\ 
$\textnormal{CCD}+\textnormal{DA}$  &   136.14     \\
\bottomrule
\end{tabular}
\label{tab:ccd_da}
} 
\end{minipage}
\hspace{0.5em}
\begin{minipage}[c]{0.25\textwidth}
\subfloat[]{
\centering
\setlength{\tabcolsep}{3.0mm}
\begin{tabular}[t]{cc}
\toprule
    TA& FVD$\downarrow$  \\	\midrule
\xmark    &     136.14   \\ 
\cmark & 118.89     \\ 
\bottomrule
\end{tabular}
\label{tab:trajectory_alignment}
}
\end{minipage}
\hspace{0.5em}
\begin{minipage}[c]{0.32\textwidth}
\subfloat[]{

\centering
\setlength{\tabcolsep}{2.5mm}
\begin{tabular}[t]{cc}
\toprule
$t$ sampler & FVD$\downarrow$ \\	\midrule
$\textnormal{uniform}(0,1)$ & 140.36 \\
$\textnormal{lognorm}(-0.8,1.0)$    & 138.5 \\ 
$\textnormal{lognorm}(-0.6,1.4)$ & 136.14  \\ 
\bottomrule
\end{tabular}
\label{tab:t_sampler}
} 
\end{minipage}
\caption{\textbf{Ablation studies}. We set the sampling steps to 4. a) Effect of CCD and DA. b) Effect of TA. c) The time schedule.}
\end{table} 
 
\subsubsection{Effect of Continuous-time Consistency Distillation (CCD)}
\vspace{-2mm}
To validate continuous-time consistency distillation effectiveness, we evaluate three training strategies: direct distribution alignment (DA), discrete-time consistency distillation (DCD) followed by DA, and continuous-time consistency distillation (CCD) followed by DA.  Table \ref{tab:ccd_da} demonstrates that consistency distillation combined with distribution alignment achieves superior performance. DA alone produces excessive trajectory deviation from the original model. Furthermore, CCD outperforms DCD through more accurate ODE trajectory preservation.
\vspace{-2mm}

\subsubsection{Effect of Distribution Alignment (DA)}
\vspace{-2mm}
To validate the effectiveness of distribution alignment, we conduct ablation studies using only continuous-time consistency loss for distillation. 
As shown in Table \ref{tab:ccd_da}, DA yields superior performance by directly approximating the real data distribution, which enhances visual fidelity under few-step sampling.
\vspace{-2mm}
\subsubsection{Effect of Trajectory Alignment (TA)}
\vspace{-2mm}
Comparing distilled models with and without trajectory alignment, Table \ref{tab:trajectory_alignment} demonstrates that model refined through TA generated higher-quality videos. This improvement stems from using DPO to align the model with high-step inference trajectories, thereby enhancing low-step inference quality.
\vspace{-2mm}

\subsubsection{Time Samplers}
\vspace{-2mm}
Prior work \citep{scale_reflow} demonstrates that the distribution of timestep affects generation quality. We train models using CCD with DA under different timestep distributions. As shown in Table \ref{tab:t_sampler}, the logit-normal timestep sampler achieves superior results.
\vspace{-2mm}

\subsubsection{Effect of Synthetic Preference Dataset in Trajectory Alignment}
\vspace{-2mm}

We compared two preference dataset strategies: synthetic preference datasets versus using real data as preferred samples with low-step generated data as less preferred samples. As shown in Figure \ref{fig:ablation_dpo}, synthetic preference datasets enable rapid improvement in few-step inference quality (red line), while real data as preferred samples leads to slow convergence (blue line). This occurs because synthetic preferred data implicitly contains trajectory information, allowing the model to quickly enhance low-step generation by aligning with high-step inference trajectories.
\vspace{-2mm}
\subsubsection{Effect of Reflow Loss in Trajectory Alignment}
\vspace{-2mm}
We investigated the impact of reflow loss by comparing models trained using $\mathcal{L}_{\textnormal{dpo}}$
alone versus those trained with the combined objective $\mathcal{L}_{\textnormal{DPO}} + \lambda_{\textnormal{RF}}\cdot\mathcal{L}_{\textnormal{RF}}$. As shown in Figure \ref{fig:ablation_dpo} (red line vs. green line), incorporating reflow loss enhances peak performance and stabilizes training.

\section{Conclusions}
\vspace{-2mm}
In this paper, we propose \textbf{\emph{SwiftVideo}}, a unified distillation framework that overcomes the limitations of prior approaches. Through continuous-time consistency distillation, we maintain precise ODE trajectory preservation while eliminating discretization errors. The incorporation of direct distribution alignment on real data enhances the visual fidelity of generated videos and overcomes teacher model performance constraints. Additionally, trajectory alignment across different inference steps serves as an effective post-training strategy to improve generation quality under few-step settings. Extensive experiments on OpenVid-1M demonstrate that \textbf{\emph{SwiftVideo}} significantly outperforms state-of-the-art distillation methods on both FVD and VBench metrics. Future work will focus on extending this framework to larger-scale video models with even fewer steps.





\setcitestyle{numbers,square}
\bibliography{yoUTU_bib}

\appendix
In the appendix, we provide additional details to complement the main paper. It includes discussions with concurrent work MeanFlow \citep{meanflow}, computational details of continuous-time consistency distillation, and supplementary information on trajectory alignment. Furthermore, we present quantitative comparison results on the VBench-I2V dataset and qualitative results on external data, demonstrating the strong generalization capability of our method. Additionally, we provide  ablation studies that validate the robustness of hyperparameter tuning in trajectory alignment.
\section{Discussion with MeanFlow}
\label{appendix:A}
\vspace{-1mm}
MeanFlow model \citep{meanflow} is a concurrent work that presents a one-step image generation model trained from scratch. As introduced in the Preliminary section, the velocity modeled in Flow Matching represents the instantaneous velocity. The marginal velocity field $\boldsymbol{v}(\mathbf{x}_t,t)$ is defined as $\boldsymbol{v}(\mathbf{x}_t,t)=\frac{d\mathbf{x}_t}{dt}$, where $'$ represents the time derivative. And in rectified flow $\boldsymbol{v}(\mathbf{x}_t,t)=\mathbf{x}_{1}-\mathbf{x}_{0}$. 
Unlike the instantaneous velocity used in Flow Matching model, this work introduces the concept of average velocity to describe flow fields and leverages the identity between average and instantaneous velocities to train the one-step image generation model. In MeanFlow model, the average velocity $u$ is :
 \begin{equation}
     u(\mathbf{x}_t,r,t)\triangleq\frac{1}{t-r}\int_r^tv(\mathbf{x}_\tau,\tau)d\tau.
 \end{equation}

Following the definition of average velocity in \citep{meanflow}, $u(\mathbf{x}_t,t)$ represents the average velocity between the time steps 0 and t, so we have $u(\mathbf{x}_t,t)t=\mathbf{x}_t-\mathbf{x}_0$. Now we differentiate both sides with respect to t:
\begin{equation}
    u(\mathbf{x}_t,t)=\frac{d\mathbf{x}_t}{dt}-t\frac{du(\mathbf{x}_t,t)}{dt}
    \label{eq:average}
\end{equation}
In the Continuous-time Consistency Distillation section, we derive the training objective as $\nabla_\theta \mathrm{E}_{\mathbf{x}_t, t}\|F_\theta(\mathbf{x}_t, t)-
      F_{\theta^-}(\mathbf{x}_t, t)-\frac{\mathrm{d} f_{\theta^-}(\mathbf{x}_t, t)}{\mathrm{d}t}\|_2^2$. 
If we substitute $\frac{\mathrm{d} f_{\theta^-}(\mathbf{x}_t, t)}{\mathrm{d}t}=\frac{\mathrm{d}\mathbf{x}_t}{\mathrm{d}t}-F_{\theta^-}(\mathbf{x}_t,t)-t\frac{\mathrm{d}F_{\theta^-}(\mathbf{x}_t,t)}{\mathrm{d}t}$ into above equation and do not employ EMA updates (i.e., $\theta^{-}=\theta$), the training objective becomes $\nabla_\theta \mathrm{E}_{\mathbf{x}_t, t}\|F_\theta(\mathbf{x}_t, t)-\frac{\mathrm{d}\mathbf{x}_t}{\mathrm{d}t}+t\frac{\mathrm{d}F_{\theta^-}(\mathbf{x}_t,t)}{\mathrm{d}t}\|^{2}$.
If we use the deep neural network $F_{\theta}(\mathbf{x}_t,t)$ to predict the average velocity $u(\mathbf{x}_t,t)$ in Eq (\ref{eq:average}), we find that this identity is formally identical to the training objective in continuous-time consistency distillation. Since Meanflow is trained from scratch, $\frac{d\mathbf{x}_t}{dt}=\mathbf{x}_1-\mathbf{x}_{0}$, whereas in our approach, where we perform distillation from a teacher model, $\frac{d\mathbf{x}_t}{dt}=F_{\textnormal{pretrain}}(\mathbf{x}_t,t)$

This finding indicates that the training objective of the continuous-time consistency model is to enable the neural network $F_{\theta}(\mathbf{x}_t,t)$ to learn to predict the average velocity between two time steps, 0 and $t$. Since current state-of-the-art video generation models are typically based on flow matching, where the neural network predicts the instantaneous velocity at time step $t$, the training process of the continuous-time consistency model inherently involves a transition from instantaneous velocity to average velocity. 

\subsection{JVP Computation Detail}
In continuous-time consistency distillation, we need to compute $\frac{\mathrm{d}F_{\theta^-}(\mathbf{x}_t,t)}{\mathrm{d}t}$ via the Jacobian-vector product (JVP), where the expression is given as follows:
\begin{equation}
    \frac{\mathrm{d}F_{\theta^-}(\mathbf{x}_t,t)}{\mathrm{d}t}=\nabla_{\mathbf{x}_t}F_{\theta^-}(\mathbf{x}_t,t)\frac{\mathrm{d}\mathbf{x}_t}{\mathrm{d}t}+\partial_tF_{\theta^-}(\mathbf{x}_t,t)
    \label{eq:jvp}
\end{equation}
This equation demonstrates that the total derivative is obtained from the JVP between $[\partial_{x}F_{\theta^-},\partial_{t}F_{\theta^{-}}]$ (the Jacobian matrix of the function $F_{\theta^{-}}$) and the tangent vector $[\frac{d\mathbf{x}_t}{dt}, 1]$. Additionally, we apply a stop-gradient operation to $\frac{\mathrm{d}F_{\theta^-}(\mathbf{x}_t,t)}{\mathrm{d}t}$, which eliminates the need for backpropagation through this term during JVP computation, thereby improving computational efficiency. We utilize $\textnormal{torch.autograd.functional.jvp}$ in PyTorch to compute the JVP. Consequently, Eq (\ref{eq:jvp}) can be computed as $\textnormal{jvp}(F_{\theta^{-}}, (\mathbf{x}_t, t), (\frac{d\mathbf{x}_t}{dt}, 1))$.

Since contemporary video generation models such as Wan 2.1 typically comprise multiple DiT blocks, computing JVP directly on the entire model leads to memory overflow issues. To address this limitation, we design an iterative algorithm that computes JVP block-by-block to control memory consumption. Specifically, we use the JVP computation output from the previous DiT block as input for the subsequent DiT block's JVP computation, while detaching intermediate results from the computational graph, thereby effectively controlling memory usage. Algorithm \ref{alg:block_jvp} presents the detailed procedure of our block-wise JVP computation.

\begin{algorithm}[tb]
\caption{Block-wise JVP Computation}
\label{alg:block_jvp}
\begin{algorithmic}[1]
\STATE \textbf{Input}: Input $\mathbf{x}_t$, time $t$, tangent vectors $(\mathbf{v}_x, \mathbf{v}_t)$, DiT blocks $\{\mathcal{B}_i\}_{i=1}^N$, time embedding $\mathcal{T}$, patch embedding $\mathcal{P}$, output layer $\mathcal{O}$ \\
\STATE \textbf{Output}: $\bm{F}_{\theta^-}(\mathbf{x}_t, t)$ and $\frac{\mathrm{d}\bm{F}_{\theta^-}(\mathbf{x}_t, t)}{\mathrm{d}t}$

\STATE // Time embedding JVP
\STATE $\mathbf{t}_{\textnormal{emb}}, \mathbf{v}_{t}^{\textnormal{curr}} \gets \texttt{jvp}(\mathcal{T}, t, \mathbf{v}_t)$

\STATE // Patch embedding JVP  
\STATE $\mathbf{h}_{\textnormal{curr}}, \mathbf{v}_{x}^{\textnormal{curr}} \gets \texttt{jvp}(\mathcal{P}, \mathbf{x}_t, \mathbf{v}_x)$
\STATE $\mathbf{g}_{\textnormal{curr}} \gets \mathbf{v}_{x}^{\textnormal{curr}}$

\FOR{$i = 1$ \TO $N$}
    \STATE // Block-wise JVP computation
    \STATE $\mathbf{h}_{\textnormal{next}}, \mathbf{g}_{\textnormal{next}} \gets \texttt{jvp}(\mathcal{B}_i, (\mathbf{h}_{\textnormal{curr}}, \mathbf{t}_{\textnormal{emb}}), (\mathbf{g}_{\textnormal{curr}}, \mathbf{v}_{t}^{\textnormal{curr}}))$
    \STATE $\mathbf{h}_{\textnormal{curr}} \gets \mathbf{sg}(\mathbf{h}_{\textnormal{next}})$ 
    \STATE $\mathbf{g}_{\textnormal{curr}} \gets \mathbf{sg}(\mathbf{g}_{\textnormal{next}})$ 
\ENDFOR
\STATE $\mathbf{y}, \frac{\mathrm{d}\mathbf{y}}{\mathrm{d}t} \gets \texttt{jvp}(\mathcal{O}, (\mathbf{h}_{\textnormal{curr}}, \mathbf{t}_{\textnormal{emb}}), (\mathbf{g}_{\textnormal{curr}}, \mathbf{v}_{t}^{\textnormal{curr}}))$

\RETURN $\mathbf{y}, \frac{\mathrm{d}\mathbf{y}}{\mathrm{d}t}$
\end{algorithmic}
\end{algorithm}

\begin{figure*}[t]
    \centering
    \begin{subfigure}[t]{0.48\textwidth}
        \centering
        \includegraphics[width=\linewidth]{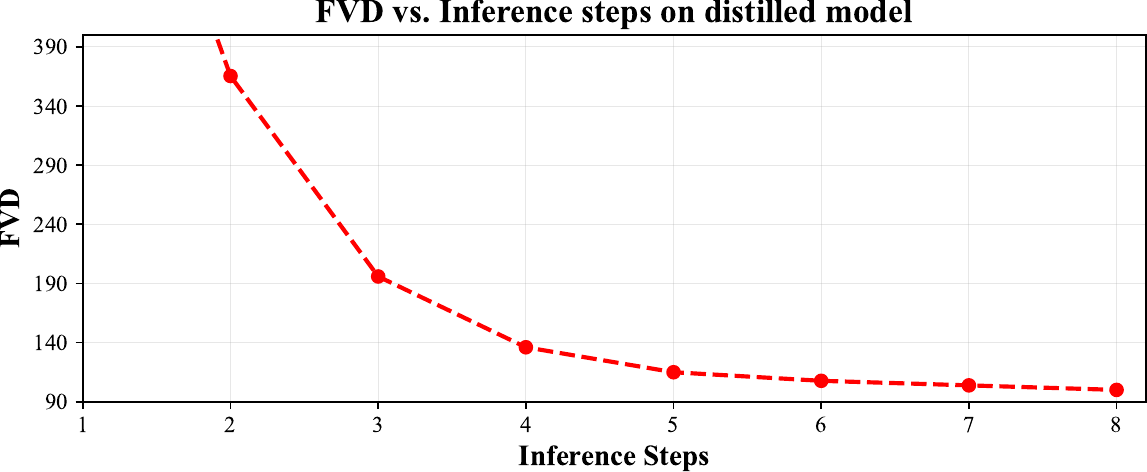}
        \caption{FVD \citep{fvd} evaluation of the distilled model using Euler solver inference on OpenVid-1M test set \citep{openvid}. The model shows significant quality improvements with increasing inference steps, particularly in the low-step regime ($\textnormal{steps}\leq8$).}
        \label{fig:inference_step}
    \end{subfigure}
    \hfill
    \begin{subfigure}[t]{0.48\textwidth}
        \centering
        \includegraphics[width=\linewidth]{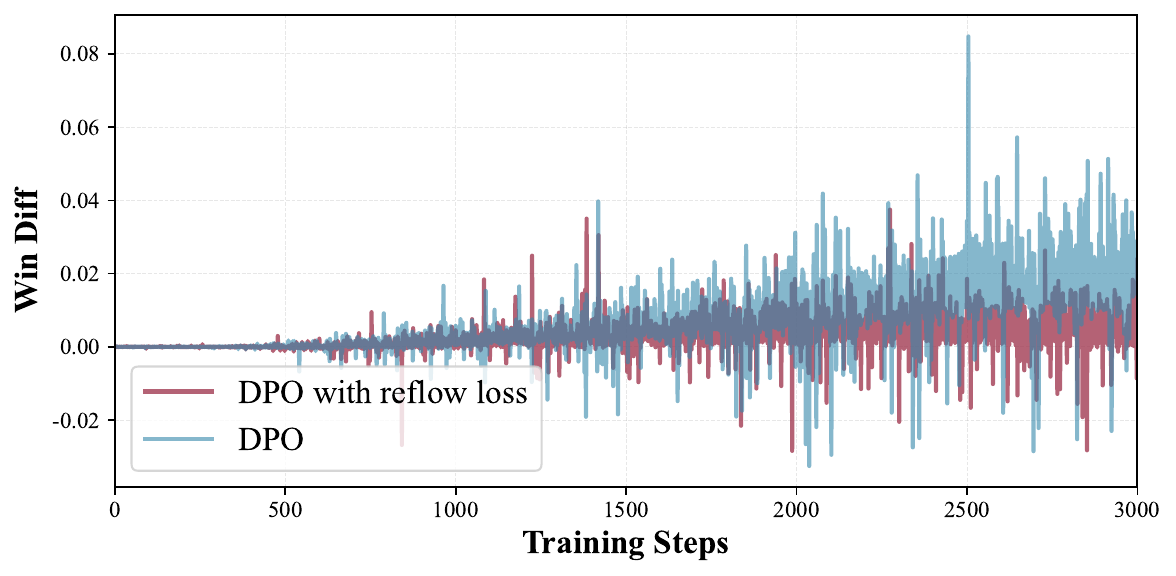}
        \caption{Win Diff during training for DPO with and without reflow loss regularization. The DPO with reflow loss (red line) effectively stabilizes the Win Diff compared to standard DPO (blue line).}
        \label{fig:win_diff}
    \end{subfigure}
\end{figure*}

\section{Details of Trajectory Alignment}
We analyzed the performance of distribution-aligned distilled models using Euler solver inference at different sampling steps. As shown in Figure \ref{fig:inference_step}, the distilled model exhibits significant quality improvements with increasing step counts in the low-step regime ($\textnormal{steps}\leq8$). This empirical observation motivates our subsequent employment of the Direct Preference Optimization (DPO) algorithm \citep{dpo} to align the model with high-step trajectories.

Additionally, we analyzed the undesired training dynamics when employing DPO loss alone for trajectory alignment. To evaluate model performance relative to the reference model, we define the win difference (Win Diff) as $\|\mathbf{v}^w-\mathbf{v}_\theta(\mathbf{x}_t^w,t)\|^2-\|\mathbf{v}^w-\mathbf{v}_\mathrm{ref}(\mathbf{x}_t^w,t)\|^2$, which measures the difference in velocity prediction errors between the training model and reference model on preferred data. As shown in Figure \ref{fig:win_diff} (blue line), training with only DPO loss leads to a continuous increase in Win Diff throughout training iterations, indicating that the model diverges from the intended trajectory and performs worse than the reference model on preferred samples. This occurs because DPO loss can decrease even when the model's velocity prediction accuracy degrades more on dispreferred data than on preferred data, resulting in suboptimal trajectory alignment. To address this issue, we introduce an additional reflow loss on preferred data as regularization to preserve high prediction fidelity. As shown in Figure \ref{fig:win_diff} (red line), incorporating the reflow loss as a regularization term effectively prevents the Win Diff from increasing during training, enabling the model to better align with preferred trajectories while diverging from dispreferred ones.

\section{More Quantitative Results}
In addition to evaluation on the Openvid-1M test set \citep{openvid}, we conduct comparative analysis against other existing distillation methods on the VBench-I2V \citep{vbench} dataset. Following the experimental setup in the main text, we generate videos at $832\times480$ resolution and 24 FPS during inference. And all experiments in this paper employ a fixed CFG \citep{cfg} scale of 3.0.

As demonstrated in Table \ref{tab:vbench}, our method significantly outperforms other existing state-of-the-art distillation methods across all quantitative metrics for both 4-step and 2-step inference. Specifically, regarding frame quality, our 4-step approach achieves results that closely approximate the teacher model's 25-step performance, while our 2-step method even surpasses the 4-step results of LCM \citep{lcm}, PCM \citep{pcm}, and DMD2 \citep{dmd2}. Furthermore, our method exceeds the teacher model's 25-step performance on image-conditioning scores. These results demonstrate that we achieve state-of-the-art performance on the VBench-I2V dataset, effectively showcasing the strong generalization capability of our approach.

\begin{table}[t]
\centering
  \begin{tabular}{c|c|cc|cc}
    \toprule
    Method & Step & \makecell{Temporal\\Quality} & \makecell{Frame\\Quality} &  \makecell{I2V\\Subject} & \makecell{I2V\\Background} \\
     \midrule
    \multirow{3}{*}{Teacher} 
    & 25 & 86.48 & 68.22  & 98.91 & 98.16\\
    & 8 & 85.2 & 67.25 & 98.79 & 98.16\\
    & 4 & 83.89 & 64.43 & 98.63 & 98.08\\
    \midrule
    LCM & 4 & 84.06 & 65.05 & 98.69 & 98.11 \\
    LCM & 2 & 81.89 & 59.09 & 98.38 & 97.93 \\
    \midrule
    PCM	& 4 & 83.92& 64.85 & 98.69 & 98.07 \\
    \midrule
    DMD2 & 4 & 84.21 &	65.32 & 98.74 & 98.09 \\
    \midrule
    $\textnormal{OSV}^\dagger$ & 4 & 84.37 &66.41 & 98.84 & 98.10\\
    $\textnormal{OSV}^\dagger$ & 2 & 82.52 & 61.11 & 98.54 & 97.95 \\
   \midrule
    \textbf{Ours} & 4 & \textbf{85.54} & \textbf{67.82}  & \textbf{99.01} & \textbf{98.17} \\
    \textbf{Ours} & 2 & \textbf{84.02} & \textbf{65.63} & \textbf{98.79} & \textbf{98.11}\\
    \bottomrule
    \end{tabular}
\caption{Image-to-Video quantitative results on the VBench-I2V dataset. $\dagger$ denotes our implementations.}
\label{tab:vbench}
\end{table}

\section{More Qualitative Results}
\begin{figure*}[t]
    \centering
    \includegraphics[width=0.85\linewidth]{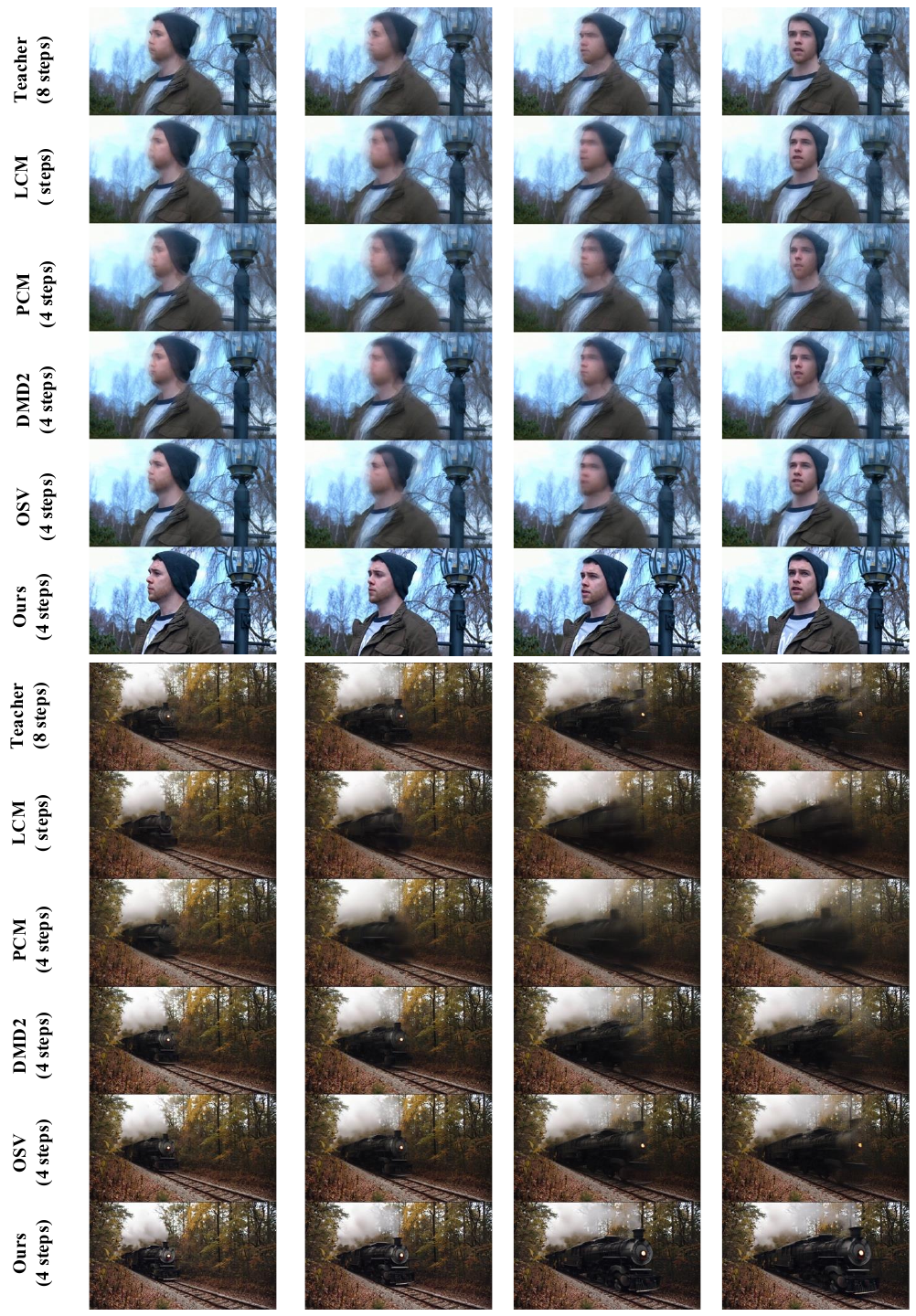}
    \caption{Qualitative comparison results of different methods on external data.}
    \label{fig:supp_cmp}
\end{figure*}

\begin{figure*}[t]
    \centering
    \includegraphics[width=0.95\linewidth]{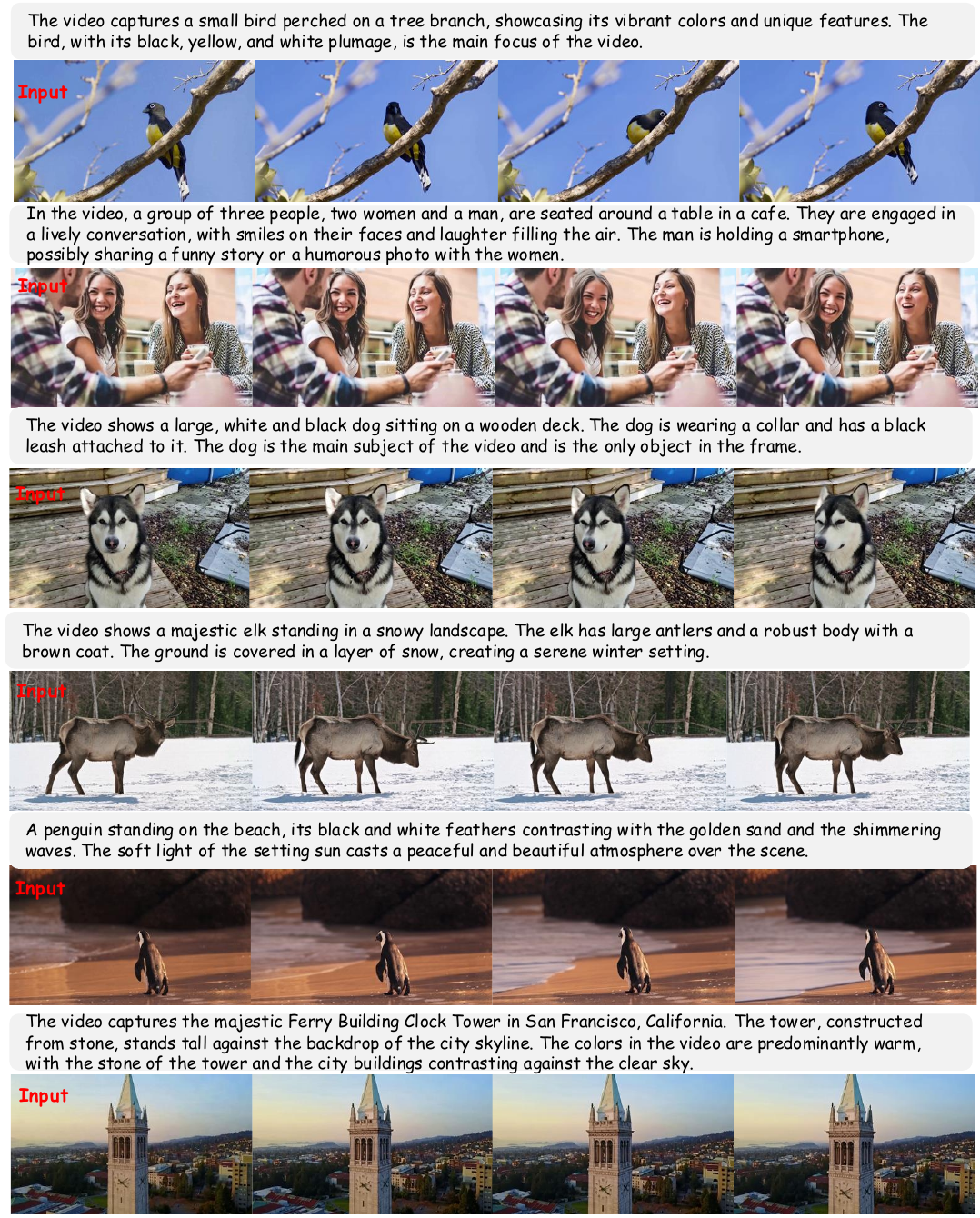}
    \caption{Qualitative visualization results on image-to-video generation. For each example, the leftmost image serves as the input condition along with the corresponding text prompt. All generated videos are synthesized using 4-step inference.}
    \label{fig:supp_vis}
\end{figure*}

We further conduct qualitative comparisons on external data. As illustrated in Figure \ref{fig:supp_cmp}, our method's 4-step inference results outperform both other distillation methods and the teacher model's 8-step inference in visual clarity and temporal consistency. Figure \ref{fig:supp_vis} demonstrates the high-quality results achieved by our method with 4-step inference on external data, further confirming the robustness and broad applicability of our approach.
\section{More Ablation Studies}
\subsection{Effect of \texorpdfstring{$\lambda_{RF}$} in Trajectory Alignment}
\begin{wrapfigure}{l}{0.6\linewidth} 
\vspace{-2mm}
    \centering
    \includegraphics[width=0.95\linewidth]{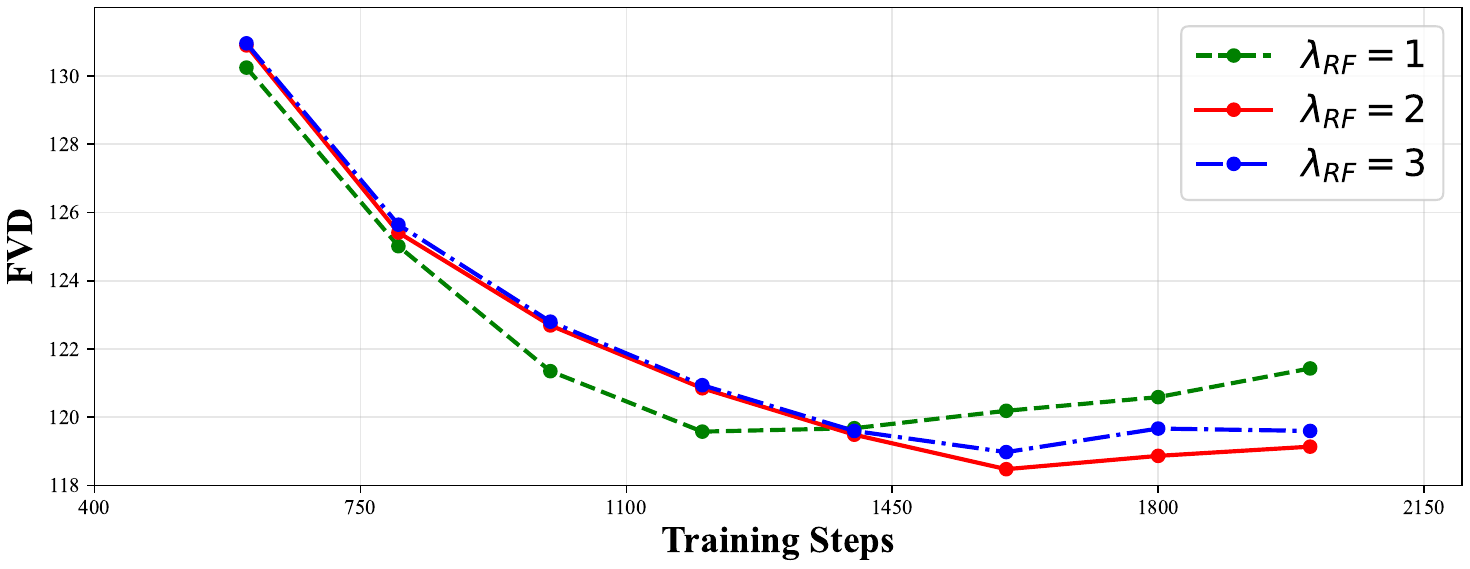}
    \caption{Effect of $\lambda_{RF}$ in Trajectory Alignment.}
    \label{fig:ablation_rf}
\end{wrapfigure}
In the main paper, we mentioned combining DPO loss with Reflow loss $\mathcal{L}_{\textnormal{DPO}} + \lambda_{\textnormal{RF}}\cdot\mathcal{L}_{\textnormal{RF}}$ to stabilize the training process of trajectory alignment. Here, we set $\lambda_{\textnormal{RF}}$ to 1, 2, and 3 to investigate the influence of $\lambda_{\textnormal{RF}}$. As shown in Figure \ref{fig:ablation_rf}, we observe that smaller values of $\lambda_{\textnormal{RF}}$ lead to faster model convergence but poorer stability in later stages, while larger values of $\lambda_{\textnormal{RF}}$ result in slower convergence but greater stability in the later phases. Overall, \textbf{model performance demonstrates robustness to the choice of $\lambda_{\textnormal{RF}}$}, and the incorporation of Reflow loss consistently yields substantial improvements in training stability.

\end{document}